\documentclass[conference]{IEEEtran}
\IEEEoverridecommandlockouts
\usepackage{cite}
\usepackage{amsmath,amssymb,amsfonts}
\usepackage{algorithmic}
\usepackage{graphicx}
\usepackage{textcomp}
\usepackage{xcolor}
\usepackage{multirow}
\def\BibTeX{{\rm B\kern-.05em{\sc i\kern-.025em b}\kern-.08em
    T\kern-.1667em\lower.7ex\hbox{E}\kern-.125emX}}

\begin{document}

\title{Automotive Radar Data Acquisition using Object Detection
}

\author{Madhumitha Sakthi\textsuperscript{1}, Ahmed Tewfik\textsuperscript{1}\thanks{1: Department of ECE, The University of Texas at Austin, United States, {madhumithasakthi.iyer@utexas.edu, tewfik@austin.utexas.edu }}
}


\maketitle

\begin{abstract}
The growing urban complexity demands an efficient algorithm to acquire and process various sensor information from autonomous vehicles. In this paper, we introduce an algorithm to utilize object detection results from the image to adaptively sample and acquire radar data using Compressed Sensing (CS). This novel algorithm is motivated by the hypothesis that with a limited sampling budget, allocating more sampling budget to areas with the object as opposed to a uniform sampling ultimately improves relevant object detection performance. We improve detection performance by dynamically allocating a lower sampling rate to objects such as buses than pedestrians leading to better reconstruction than baseline across areas with objects of interest. We automate the sampling rate allocation using linear programming and show significant time savings while reducing the radar block size by a factor of 2. We also analyze a Binary Permuted Diagonal measurement matrix for radar acquisition which is hardware-efficient and show its performance is similar to Gaussian and Binary Permuted Block Diagonal matrix. Our experiments on the Oxford radar dataset show an effective reconstruction of objects of interest with 10\% sampling rate. Finally, we develop a transformer-based 2D object detection network using the NuScenes radar and image data.
\end{abstract}


\section{Introduction}
In an autonomous driving scenario, it is necessary to acquire various sensor data and process them efficiently to get a complete perspective about the surroundings. Radar data in addition to images has aided in improving the object detection performance \cite{radar1} \cite{radar2}. During onboard radar signal acquisition, CS is a technique used for recovering the original radar data with sampling rates much lower than the Nyquist rate \cite{radar-cs2} \cite{radar-cs3}. The authors in \cite{radar-cs1}, showed successful radar reconstruction with 40\% sampling rate. 
In our method, we show efficient reconstruction with 10\% sampling rate.  
In adaptive CS, the CS technique is extended to focus on the important regions by allocating more sampling budget based on prior information \cite{adaptive-radar-cs1} \cite{adaptive-radar-cs3}. In \cite{adaptive-radar-cs2} they used adaptive compressed sensing for a static tracker case and have shown improved target tracking performance. However, in our algorithm, we used adaptive compressed sensing for radar acquisition from an autonomous vehicle where, both the vehicle and the objects were moving. 

In our work, we develop an adaptive block CS algorithm to acquire radar data using the object detection results from the image. The images were processed using Faster R-CNN \cite{fasterRCNN} network to obtain the bounding box and the class of an object. The bounding boxes in the camera coordinates were converted to bird-eye view coordinates and the azimuth of the object was identified. Therefore, the azimuth block with an object was sampled at a higher sampling rate compared to other regions which helped in the efficient reconstruction of objects such as pedestrians, bicycles, cars, trucks and buses while they were either missed or reconstructed poorly by the baseline, standard CS technique. 
In this work, we compare our method against the standard CS and another simpler algorithm called Algo-1. The standard CS acquisition has a uniform sampling rate of 10\% across all the radar frames. In Algo-1, we use the prior image information to identify the important blocks with any object of interest but, the sampling rate is allocated manually and is fixed for a block size of 50x100. In Algo-2, proposed as the main contribution in this paper, compared to Algo-1, we 1) dynamically decrease the sampling rate for car or bus compared to pedestrian and use that budget to recover the trifling areas as well apart from having better reconstruction in the important regions, 2) we automated the dynamic sampling rate allocation using Linear Programming (LP), 3) we reduced the block size of radar reconstruction by a factor of 2 to 25x100 leading to a decrease in reconstruction time by 60\% for 10\% sampling rate. Therefore with the proposed method, we could reconstruct the trifling radar regions with the same or better reconstruction quality in comparison to the baseline \& Algo-1 technique while preserving the reconstruction quality of the objects of interest. 
Therefore, our reconstruction technique \textit{could} be reliably used for radar segmentation \cite{radar-segmentation} apart from object detection task. 
In CS acquisition, it is important to design a hardware-efficient measurement matrix in order to operate the algorithm efficiently in real-time. Therefore, we analyze a Binary Permuted Diagonal (BPD) measurement matrix that simplifies hardware acquisition by directly measuring the radar data instead of acquiring a linear combination of the input radar 
and show that the reconstruction quality is similar to that of the Gaussian measurement matrix. 

The binary measurement matrices in CS have been explored by several studies. In \cite{simp-meas-matrix}, the authors proposed a Binary Permuted Block Diagonal (BPBD) matrix with equal-sized diagonal blocks permuted along the columns to create randomness. 
Binary measurement matrices have been used for image \cite{bin-image-BCS-SPL}, ECG \cite{CS-ECG-binary-meas-matrix} \cite{detr-bin-matrix-ecg} and ground-penetrating radar data acquisition \cite{binary-matrix-radar-ground}.
To the best of our knowledge, we are the first ones to implement the BPD matrix for automotive radar acquisition. Finally, we also develop an end-to-end transformer-based 2-D object detection \cite{DETR} network using the NuScenes \cite{nuscenes} radar and image dataset. 
Numerous studies \cite{radar4} \cite{rad_img1} showed the advantage of using both radar and images in an object detection network for improved object detection performance. 
\cite{radar1}, \cite{radar2_iclr} showed that radar in addition to image improved distant vehicle object detection and occluded object detection due to adverse weather conditions. 
Again, to the best of our knowledge, we are the first ones to implement a DETR based object detection network using both radar and image data and we show that the object detection performance of the model using both image and radar data performed better than the object detection model trained only on the image data.

In the next section, we explain the adaptive block CS algorithm followed by the sampling rate allocation using LP and the design of the measurement matrix along with the DETR-Radar architecture. In section 3, we present our results and finally, conclude the paper in section 4. 


\section{Method}
\subsection{Dataset}
In the Oxford radar robotcar dataset \cite{oxford}, they provide 
radar, 
rear camera data, 
front camera data among other sensors, covering 280km in Oxford, UK. The front camera was captured at 16Hz Frames per second (FPS), the rear camera at 17Hz and radar at 4 Hz. To the best of our knowledge, the Oxford dataset was the only raw radar dataset. Since manual annotation was required to identify objects 
, we tested our algorithm on three random scenes with 11 radar frames each. 
To train our DETR-Radar object detection model, we used the NuScenes v0.1 dataset \cite{nuscenes}. This is one of the publicly available datasets for autonomous driving with a range of sensors such as cameras, LiDAR and radar with 3D bounding boxes. Similar to \cite{rad_img1}, we converted all the 3D bounding box annotations to 2D bounding boxes and merged similar classes to 6 total classes, Car, Truck, Motorcycle, Person, Bus and Bicycle. This dataset consists of around 45k images and we split the dataset into 85\% training and 15\% validation.



\subsection{Adaptive block compressed sensing}
CS relies on the sparsity of the signal in some domains for the robust reconstruction of data sampled at very low sampling rates. The signal $z \in R^n$ is measured using the random measurement matrix $ \phi \in R^{mxn}$ which gives $y \in R^m$ measurements. The original signal $x$ is assumed to be sparse in Discrete Cosine Transform domain and hence, it is recovered using BP algorithm as $\min_{x} ||\theta x||_1$  $s.t.$  $\phi x = y $ with $\theta$ as the domain transformation matrix \cite{image-adaptive-cs}. In the adaptive block CS, the radar region is split into equal-sized blocks and varying measurements $m$ are allocated based on prior knowledge. 

The Oxford radar data was captured every 0.25s with a range resolution of 4.38cm and 0.9$^{\circ}$ azimuth resolution for 3768 range bins and 400 azimuth bins. Therefore, a total range of 163m and 360$^{\circ}$ horizontal field-of-view (HFOV) was captured. Also, the front camera has 66 degrees HFoV. The rear camera has 180 degrees HFoV. 
From this setup, there would be a blind spot of 57 degrees to the left and 57 degrees to the right.
In both Algo-1 and Algo-2, 
at t=0s, the images from the camera are captured and given to the Faster R-CNN object detection network. At t=0.12s, the object detection output is processed to obtain the important azimuth blocks. This data is processed to allocate more sampling budget to the important azimuth blocks and the sampling budget is used to capture the radar data at t=0.25s. This process is repeated for all the radar frames.

In Algo-1, the radar data was split into bins of size 50x100 creating 8 equal regions in azimuth and 37 regions in range. Therefore, from the camera image, the important azimuth sections to focus on would be derived based on the presence of objects in that section. Since the depth information is not available from the camera images, we can only choose azimuth sections and not the range. 
Also, since the radar data was acquired with a 163 m range, we allocated more sampling rate to the first 78.84m compared to the last 84.16m since there is not much useful information in the farther range values. The average driving speed in an urban environment is 40 miles per hour. Since radar is captured at 4 Hz, for every frame, the object could have moved 4.25m. Since the bin resolution is 4.38cm and for a particular block with 100 bins, the area spanned would be 4.38m. Since we are looking into the first 18 range blocks, that covers a total area of 78.84m. Hence, in an urban setting, the moving vehicle can be comfortably captured by focusing within the 78.84m range. In a freeway case, for an average speed of 65 miles per hour, the vehicle could have moved 7.2m per frame and again, this would be captured by focusing on the first 78.84m. Therefore, the radar regions are split into three ranges, R1, R2, R3 where, R1 is the chosen azimuth until 18\textsuperscript{th} range block (78.84m), R2, the other azimuth regions until the 18\textsuperscript{th} range block (78.84m) and finally, R3, all the azimuth blocks from 19\textsuperscript{th} to 37\textsuperscript{th} range block. Since across different scenes, there could have been a variable number of important azimuth blocks, if the chosen azimuth block was less than 50\% i.e., less than 4 out of 8, we randomly sampled more azimuth blocks. Also, if it was more than 4, we avoided losing important information by reducing the sampling rate in the R1 region. Hence, when there were 4 important azimuth blocks, R1 was sampled at 30.8\%, R2 at 5\% and R3 at 2.5\% sampling rate. In the case of 5 azimuth blocks, R1 was sampled at 25.5\%, R2 and R3 at 5\% and 2.5\% respectively. In the case of 6 azimuths, R1 at 20.2\%, R2 at 5\% and R3 at 2.5\%. Therefore, these sampling rates were set manually to determine the importance of image data as prior information for adaptive CS. 

In Algo-2, the radar data was split into bins of size 25x100, amounting to 22.5$^{\circ}$ azimuth and 4.38m range each. There were 16 blocks in azimuth and 37 blocks in range. 
The block size of 25x100 instead of 50x100 from Algo-1 reduced the processing time by 60\% for a 10\% sampling rate. This empirical change in processing time was measured using MATLAB on a 3.1 GHz Dual-Core Intel Core i7 processor. The dynamic sampling rate for Algo-2 was determined based on the importance of a region and was allocated using linear programming (LP). Where, $a_1$ corresponds to blocks with pedestrians or bicycles, $a_2$ corresponds to blocks with the car and $a_3$ represents all the other regions. 
$r_1$ is the radar region within the first 18 range blocks (78.84m) and $r_2$ is the next 19 range blocks (83.22m). $x_1, x_2, x_3, x_4$ is the sampling rate for each block, which is being optimized. $S$ is the total radar data and $0.1S$ corresponds to the 10\% sampling rate. The objective function is the total sampling budget allocated across various azimuth blocks for a given radar frame and it is optimized such that, for a given frame, the sampling rate does not exceed 10\%. Since $x1$ is the sampling rate for the region with small objects, the constraints are set such that, it is thrice as big as the least important region's sampling rate and $x2$ is twice as big. Also, $x4$ is the sampling rate for the region starting 78.84m from the autonomous vehicle and hence it is optimized to have the least sampling rate of less than 2.5\%.

For any vector $x\in \mathbb{R}^4$, let 
\vspace{-3mm}
\begin{align*}
    f(x) &= a_1r_1x_1 + a_2r_1x_2 \ldots \\
    &~~~~~+ a_3r_1x_3 + (a_1 + a_2 + a_3)r_2x_4
\end{align*}
\vspace{-3mm}
Then, we have the following linear program
\begin{align*}
\max_{x\geq 0} &\; f(x)\\
s.t. &\; x_1 - 3x_3 = 0,\; x_2 - 2x_3 = 0\\
& f(x) \leq 0.1S, \;0.02 \leq x_4 \leq 0.025,\\
& 0.05\leq x_i\leq 0.4 ~~\text{for } i = 1,2,3\\
\end{align*}
\vspace{-10mm}

Therefore, the bounding boxes and classes from the object detection network are used to determine the azimuth blocks $a_1, a_2, a_3$, given as a constant to the LP. 
The LP algorithm in one case, gave 37.39\%, 24.93\% 12.46\% and 2.5\% as $x_1,x_2,x_3,x_4$  sampling rates.

\begin{figure*}[ht!]
\begin{center}
\includegraphics[width=15.5cm]{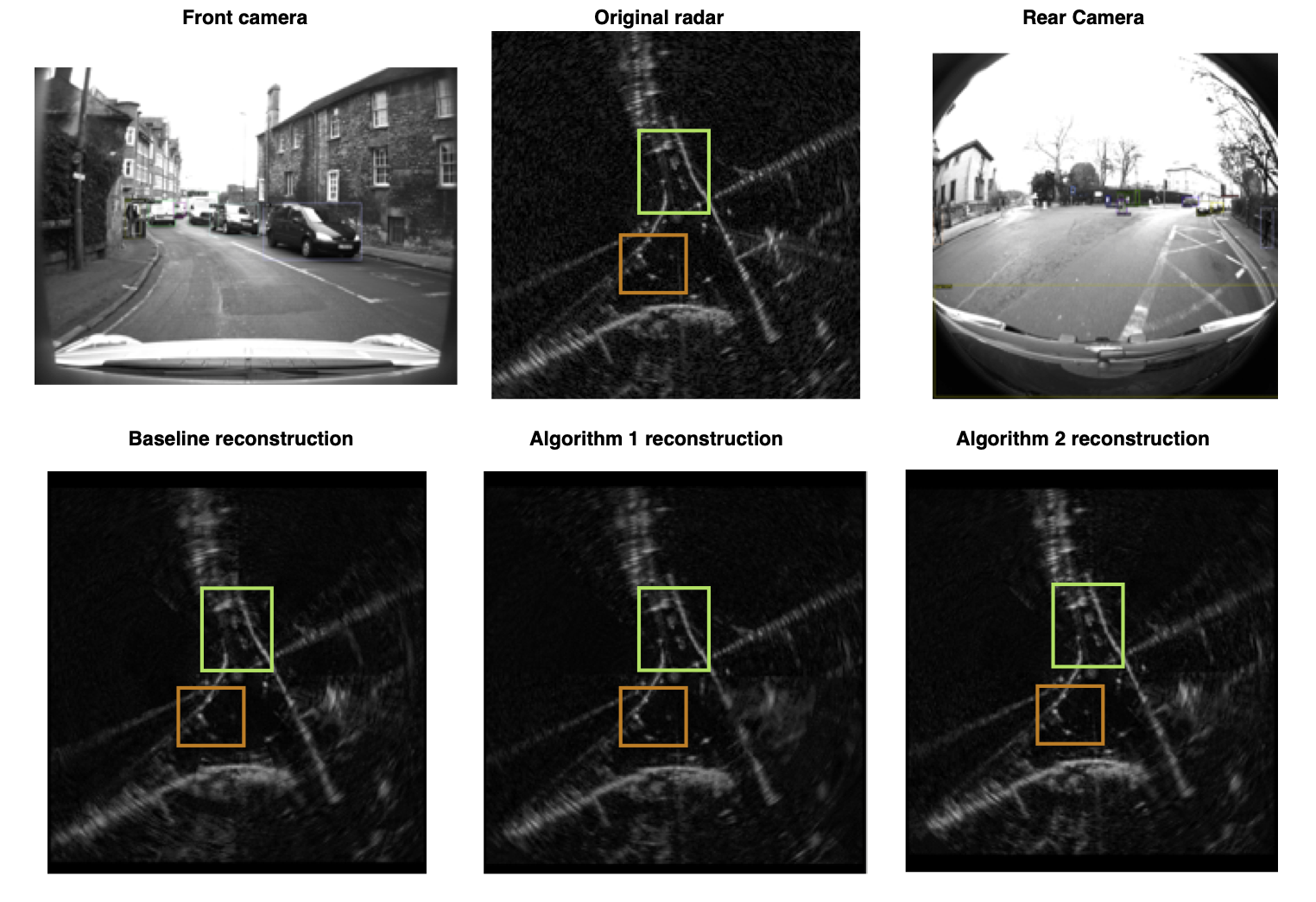}
\centering
\caption{In scene 2, frame 11, the front camera, rear camera, original radar and reconstructions are shown. The green box highlights the cars to the front and the pedestrian. The pedestrian was missed by the baseline reconstruction but, it is captured by Algorithm 1 and it is more clear in Algorithm 2. The orange box highlights the cars and bicycle to the rear. The bicycle was captured by Algorithm 2 but, it is hardly reconstructed by Algorithm 1 and baseline.}
\label{scene2-frame11}
\end{center}
\vspace{-6mm}
\end{figure*}

\subsection{Measurement matrix analysis}
We compare the standard Gaussian measurement matrix and BPBD matrix with the BPD matrix. In the case of our BPD matrix, it is designed such that all elements of a row are 0 except for one randomly selected column. Therefore, when each row of the measurement matrix is used to measure the original signal, only one value of the original data is retained, even the addition of a few measurements in the case of BPBD is avoided and addition, multiplication of values in the case of the Gaussian measurement matrix is eliminated. Therefore, complex multipliers for measurement using the Gaussian matrix which consumes higher power are avoided and replaced with simpler elements like switches and selectors for binary matrix making them hardware efficient \cite{bin-image-BCS-SPL}. 

\subsection{DETR-Radar}
The Faster R-CNN \cite{fasterRCNN} object detection network is one of the well-refined techniques for 2D object detection. However, it relies heavily on two components, non-maximum suppression and anchor generation. The end-to-end transformer-based 2D object detection introduced in \cite{DETR}, eliminates the need for these components. We included radar data in two ways. In the first case, we included the radar data as an additional channel to the image data \cite{radar1}. In the second case, we rendered the radar data on the image. We used perspective transformation based on \cite{rad_img1} to transform radar data points from the vehicle coordinates system to camera-view coordinates. 
In all the above cases, the models were pre-trained on the COCO dataset and we fine-tuned them on the NuScenes data. To the best of our knowledge, we are the first ones to implement end-to-end transformer-based 2D object detection using both image and radar data.
%

\section{Experiment}
\begin{table}[ht!]
\vspace{-4mm}
\caption{\label{results}
S-CS denotes standard CS, A-1 is algorithm 1 and A-2 is algorithm 2. The presence of an object is indicated as 'yes' and if the object is faint or absent, it is indicated as 'no'.}
\centering
\begin{tabular}{|c| c| c| c| c| c|} 
 \hline
 Scene & Frame &Object & S-CS & A-1 & A-2 \\ [0.5ex] 
 \hline\hline
 \multirow{3}{4em}{Scene1} &  5,11 & car (top)  & no & yes&  yes \\
 &  1-4 & Person (left)  & no & yes&  yes \\
 & 5 & Person(left)  & no & yes & \textbf{no}  \\
 & 2-5,8,10 & Person (top-left) & no & yes & yes \\
 & \textbf{6,7,9} & \textbf{Person (top-left)} & \textbf{no} & \textbf{no} & \textbf{yes} \\
 & 8-11 & Bicycle (top) & no & yes & yes \\
 & \textit{2,3,5} & \textit{Pole (right) } & \textit{yes} & \textit{no} & \textit{yes} \\
 \hline
 \multirow{3}{4em}{Scene2} &  1-11 & Cars (top-right)  & no & yes&  yes \\
 &  2,3,6 & Car (rear)  & no & yes&  yes \\
 & 7-9 & Bicycle (rear) & no & yes & yes \\
 & \textbf{10,11} & \textbf{Bicycle (rear)}  & \textbf{no} & \textbf{no} & \textbf{yes}\\
 &1-3 & Person (rear)  & no & yes & yes\\ 
 &5-7,9 & Person (rear-left) & no & yes & yes \\
 & 7-11 & Person(front) & no & yes &yes \\
 & \textit{7,8} & \textit{Traffic light (rear)} & \textit{yes} & \textit{no} & \textit{yes} \\
 & \textit{1} & \textit{Pole (top-right)} & \textit{yes} & \textit{no} & \textit{yes}\\
 \hline
 \multirow{2}{4em}{Scene3} &  3,5,6 & car (rear)  & no & yes & yes \\
 &  7,8 & car (top-right)  & no & yes & yes \\
 &  \textbf{4} & \textbf{Bicycle (right)} & \textbf{yes} & \textbf{no} & \textbf{yes} \\
 & \textbf{1,8} & \textbf{Person (left)} & \textbf{yes} & \textbf{no} & \textbf{yes} \\
 &10 & Person (rear-left) & no & yes & \textbf{no} \\
 & \textit{3,5,10} & \textit{Pole (right)} & \textit{yes} & \textit{no} & \textit{yes} \\
 \hline
\end{tabular}
\vspace{-4mm}
\end{table}

The algorithms were tested on three random scenes with 11 radar frames each \cite{oxford}. Since the important information could be captured within the first 78.84m in an urban setting, the results were analyzed in that region and are shown in the figure \ref{scene2-frame11} as well. The standard CS refers to the baseline CS algorithm, with a uniform sampling rate of 10\% on the entire radar frame (163m, 360$^{\circ}$ azimuth). The Algo-1 is a basic version with 50x100 block size and manual sampling rate allocation and Algo-2 is the promoted algorithm with 25x100 block size and LP-based dynamic sampling rate allocation. All the reconstruction algorithms have BPD as the measurement matrix. In table \ref{results}, the presence of an object is indicated as 'yes' and if it is absent or faint, it is indicated as 'no'. In the first scene, 
the person in the top-left was missed by both standard CS and poorly reconstructed by Algo-1 in frame 6,7 and 9. However, it was recovered by Algo-2 because it had a significantly higher sampling rate allocated to regions with pedestrians. However, the person to the left of the vehicle appeared in the blind spot of the camera and it was missed by our Algo-2. But, it was captured by Algo-1 because the block adjacent to it was selected and it was twice as big as the second algorithm's block size. 
Therefore, by using additional camera information, in Algo-2, the person could be captured. However, Algo-2 in general has better or similar reconstruction across all the regions of $r_1$ compared to baseline and it captured the pole to the right in frames 2,3 and 5.
In scene 2,
the bicycle to the rear was missed by the baseline and poorly reconstructed by Algo-1. However, it was reconstructed by Algo-2 in frames 10 and 11. The traffic light to the rear in frames 7, 8 and pole in frames 1 to the top-right was missed by Algo-1 but, it was captured by Algo-2. 
In scene 3 
frame 4, the bicycle to the right was poorly reconstructed by Algo-1 but it was captured by Algo-2. The person to the left in frame 1 and 8 was missed by Algo-1 but it was captured by Algo-2. In Algo-1, apart from the important regions, the sampling rate was drastically reduced in other regions. The right and left of the vehicle is in the blind spot of the camera and would be classified as non-important regions. However, in Algo-2, the sampling rates saved from regions with bus or car were used to redistribute it across other regions and this helped in capturing the objects to the left and right. The person to the rear-left in frame 10 was missed by Algo-2 but it was captured by Algo-1. Although the person was captured by the object detection network, it was missed by our algorithm.
Similar to the previous scenes, the pole to the right was captured by Algo-2 in frames 3,5 and 10 while it was missed by Algo-1. Also, from table \ref{results}, considering ‘yes’ as detected and ‘no’ as not detected, Algo-1 detected 52 out of 60 objects across frames and scenes. Whereas, the improved Algo-2, detected 58 out of 60 objects, improving from 86\% detection using algorithm1 to 96\% detection rate using our proposed Algo-2. Therefore, the Algo-2 has led to significant improvement in better reconstruction/detection.

In table \ref{meas-matrix-result}, the average peak signal-to-noise ratio (PSNR) is reported across three measurement matrices.
In general, the BPD matrix has slightly lower PSNR compared to Gaussian measurement matrices. However, BPD is hardware efficient since it has binary elements as the measurement matrix. 
\begin{table}[ht!]
\caption{\label{meas-matrix-result} The three scenes were reconstructed using the standard CS algorithm and average PSNR is reported.}
\centering
\begin{tabular}{|c| c| c| c| c|} 
 \hline
 Sampling rate & Scene & Gaussian &  BPBD & BPD \\ [0.5ex] 
 \hline\hline
 \multirow{3}{4em}{10\%} & scene1 &29.88 &29.91 &29.87 \\
 & scene2 &30.18 &30.22 &30.17 \\
 & scene3 &29.89 &29.93 &29.88 \\
 \hline
 \multirow{3}{4em}{20\%} &   scene1 &32.00 &32.00 &32.00 \\
 & scene2 &32.26 &32.26 &32.26 \\
 & scene3 &32.08 &32.09 &32.07 \\
 \hline
 \multirow{2}{4em}{30\%} &   scene1 &34.30 &34.30 &34.27 \\
 & scene2 &34.53 &34.52 &34.49 \\
 & scene3 &34.43 &34.42 &34.38 \\
 \hline
\end{tabular}
\vspace{-2mm}
\end{table}

\begin{table}[h]
\caption{\label{DETR-results} I denote model trained on Images, I+R indicated model trained with radar as an additional channel and RonI is for a model trained with the radar rendered on the image. F-R is the Fast R-CNN network and FA-R is the Faster R-CNN network.}
\begin{center}
\begin{tabular}{|c| c| c| c| c| c| c| c|} 
 \hline
 Network & AP &  AP50 & AP75 & AR & ARs & ARm & ARl \\ [0.5ex] 
 \hline\hline
 F-R \cite{rad_img1} & 35.5 & 59.0 & 37.0 & 42.1 &21.1 &39.1 &51.4 \\
\hline
 FA-R(I) &  39.5 &67.8 &41.7 &47.0 &25.6 &44.4 &56.8 \\
 \hline
 FA-R(I+R) & 46.2 &73.8 &50.3 &53.0 &32.8 &51.5 &59.9 \\
 \hline
 FA-R(RonI) &  38.0 &65.4 &40.0 &44.9 &17.6 &42.1 &56.3 \\
 \hline
 DETR(I) &   47.1 &80.2 &50.4 &61.6 &38.4 &57.2 &72.5\\
 \hline
 DETR(RonI) &  48.6 & 80.4 & \textbf{52.7} & \textbf{63.6} & 40.1 & \textbf{60.2} & 73.1\\
 \hline
 DETR(I+R) &  44.8 &76.3 &46.8 &58.2 &29.7 &54.9 &68.8 \\
 \hline
\end{tabular}
\end{center}
\vspace{-5mm}
\end{table}

Finally, we trained a separate object detection network using the NuScenes image and radar data. In this case, we limited our analysis to NuScenes image and original radar data because the Nuscenes radar that was available to us were processed pointclouds with annotations. Whereas, the Oxford data was the only available raw data on which we could apply CS but, without object annotations. Hence, we could not use Oxford data to train the object detection model. 
As shown in the table \ref{DETR-results}, our baseline comparison is with the \cite{rad_img1} paper, where, they trained the model on Nuscenes v0.1 image dataset and used the radar data for anchor generation. The Faster R-CNN Img and Faster R-CNN RonImg models had ResNet-101 \cite{ResNet} as the backbone structure \cite{Detectron2018}. 
The models with I+R were trained with radar as an additional channel. Therefore, the first layer of the backbone structure was changed to process the additional radar channel. 
The DETR network \cite{DETR} had ResNet-50 \cite{ResNet} as the backbone structure.
The I and RonI models were trained for the same number of epochs for a fair comparison. The I+R models were trained for additional epochs since the backbone structure's first layer was modified. 
In the Faster R-CNN case, I+R has better performance than I. While, in DETR, RonI has better performance. 
The Faster R-CNN I and RonI were trained for 25k iterations. The Faster R-CNN I+R was trained for 125k iterations. DETR I and DETR RonI models were trained for 160 epochs. While DETR I+R was trained for 166 epochs. The DETR RonI model performed better across various metrics compared to the baseline, Faster R-CNN and DETR I+R model. We believe that the attention heads in the transformer architecture helped in focusing object detection predictions around the radar points. However, the Faster R-CNN I+R was better than the Faster R-CNN RonI model. We used the standard evaluation metrics, mean average precision (AP), mean average recall (AR), average precision at 0.5, 0.75 IOU, small, medium and large AR \cite{coco-eval}.

\section{Conclusion}
In this paper, we propose an adaptive block CS for radar data acquisition using an object detection result as prior information. Therefore, the main contribution of our work is to determine the areas with the object in the radar frame using prior image information while processing the data in a real-time scenario. The algorithm dynamically saves on sampling rate from big objects
and redistributes to other regions to efficiently capture poles and other objects. 
This algorithm uses LP for dynamic sampling rate allocation and the reduction of block size helps in reducing the processing time significantly. Also, a hardware-efficient BPD measurement matrix is compared with a standard measurement matrix. Finally, our end-to-end transformer-based model trained on image and radar has better object detection performance than Faster R-CNN and transformer-based model trained on just images, validating the necessity for radar in addition to images.

\bibliographystyle{template}
\bibliography{template.bib}

\begin{thebibliography}{10}

\bibitem{radar1}
M.~{Meyer} and G.~{Kuschk},
\newblock ``Deep learning based 3d object detection for automotive radar and
  camera,''
\newblock in {\em 2019 16th European Radar Conference (EuRAD)}, 2019, pp.
  133--136.

\bibitem{radar2}
Shuo Chang and Yifan et~al. Zhang,
\newblock ``Spatial attention fusion for obstacle detection using mmwave radar
  and vision sensor,''
\newblock {\em Sensors (Basel, Switzerland)}, vol. 20, no. 4, February 2020.

\bibitem{radar-cs2}
Z.~{Slavik} and A.~{Viehl} et~al.,
\newblock ``Compressive sensing-based noise radar for automotive
  applications,''
\newblock in {\em 2016 12th IEEE International Symposium on Electronics and
  Telecommunications (ISETC)}, 2016, pp. 17--20.

\bibitem{radar-cs3}
A.~{Correas-Serrano} and M.~A. {González-Huici},
\newblock ``Experimental evaluation of compressive sensing for doa estimation
  in automotive radar,''
\newblock in {\em 2018 19th International Radar Symposium (IRS)}, 2018, pp.
  1--10.

\bibitem{radar-cs1}
F.~{Roos} and P.~{Hügler} et~al.,
\newblock ``Data rate reduction for chirp-sequence based automotive radars
  using compressed sensing,''
\newblock in {\em 2018 11th German Microwave Conference (GeMiC)}, 2018, pp.
  347--350.

\bibitem{adaptive-radar-cs1}
A.~M. {Assem}, R.~M. {Dansereau}, and F.~M. {Ahmed},
\newblock ``Adaptive sub-nyquist sampling based on haar wavelet and compressive
  sensing in pulsed radar,''
\newblock in {\em 2016 4th International Workshop on Compressed Sensing Theory
  and its Applications to Radar, Sonar and Remote Sensing (CoSeRa)}, 2016, pp.
  173--177.

\bibitem{adaptive-radar-cs3}
J.~{Zhang}, D.~{Zhu}, and G.~{Zhang},
\newblock ``Adaptive compressed sensing radar oriented toward cognitive
  detection in dynamic sparse target scene,''
\newblock {\em IEEE Transactions on Signal Processing}, vol. 60, no. 4, 2012.

\bibitem{adaptive-radar-cs2}
I.~{Kyriakides},
\newblock ``Adaptive compressive sensing and processing for radar tracking,''
\newblock in {\em 2011 IEEE International Conference on Acoustics, Speech and
  Signal Processing (ICASSP)}, 2011, pp. 3888--3891.

\bibitem{fasterRCNN}
Shaoqing Ren and Kaiming~He et~al.,
\newblock ``Faster r-cnn: Towards real-time object detection with region
  proposal networks,'' 2015.

\bibitem{radar-segmentation}
Prannay Kaul and Daniele De~Martini et~al.,
\newblock ``Rss-net: Weakly-supervised multi-class semantic segmentation with
  fmcw radar,'' 2020.

\bibitem{simp-meas-matrix}
Z.~{He}, T.~{Ogawa}, and M.~{Haseyama},
\newblock ``The simplest measurement matrix for compressed sensing of natural
  images,''
\newblock in {\em 2010 IEEE International Conference on Image Processing},
  2010, pp. 4301--4304.

\bibitem{bin-image-BCS-SPL}
A.~{Akbari} and M.~{Trocan},
\newblock ``Robust image reconstruction for block-based compressed sensing
  using a binary measurement matrix,''
\newblock in {\em 2018 25th IEEE International Conference on Image Processing
  (ICIP)}, 2018.

\bibitem{CS-ECG-binary-meas-matrix}
Ren Ren and Xianxiang et~al. Chen,
\newblock ``Compressed sensing based method for electrocardiogram monitoring on
  wireless body sensor using binary matrix,''
\newblock {\em International Journal of Wireless and Mobile Computing}, vol. 8,
  no. 2, pp. 114--121, 2015.

\bibitem{detr-bin-matrix-ecg}
A.~{Ravelomanantsoa}, H.~{Rabah}, and A.~{Rouane},
\newblock ``Compressed sensing: A simple deterministic measurement matrix and a
  fast recovery algorithm,''
\newblock {\em IEEE Transactions on Instrumentation and Measurement}, vol. 64,
  no. 12, pp. 3405--3413, 2015.

\bibitem{binary-matrix-radar-ground}
L.~{Miccinesi} and N.~{Rojhani} et~al.,
\newblock ``Compressive sensing for no-contact 3d ground penetrating radar,''
\newblock in {\em 2018 41st International Conference on Telecommunications and
  Signal Processing (TSP)}, 2018, pp. 1--5.

\bibitem{DETR}
Nicolas Carion and Francisco~Massa et~al.,
\newblock ``End-to-end object detection with transformers,'' 2020.

\bibitem{nuscenes}
Holger Caesar and Varun~Bankiti et~al.,
\newblock ``nuscenes: A multimodal dataset for autonomous driving,'' 2020.

\bibitem{radar4}
Shuo Chang and Yifan et~al. Zhang,
\newblock ``Spatial attention fusion for obstacle detection using mmwave radar
  and vision sensor,''
\newblock {\em Sensors (Basel, Switzerland)}, vol. 20, no. 4, February 2020.

\bibitem{rad_img1}
R.~{Nabati} and H.~{Qi},
\newblock ``Rrpn: Radar region proposal network for object detection in
  autonomous vehicles,''
\newblock in {\em 2019 IEEE International Conference on Image Processing
  (ICIP)}, 2019, pp. 3093--3097.

\bibitem{radar2_iclr}
S.~{Chadwick}, W.~{Maddern}, and P.~{Newman},
\newblock ``Distant vehicle detection using radar and vision,''
\newblock in {\em 2019 International Conference on Robotics and Automation
  (ICRA)}, 2019, pp. 8311--8317.

\bibitem{oxford}
Dan Barnes and Matthew~Gadd et~al.,
\newblock ``The oxford radar robotcar dataset: A radar extension to the oxford
  robotcar dataset,'' 2019.

\bibitem{image-adaptive-cs}
Irfan Mehmood and Ran et~al. Li,
\newblock ``Adaptive compressive sensing of images using spatial entropy,''
\newblock {\em Computational Intelligence and Neuroscience)}, 2017.

\bibitem{ResNet}
Kaiming He and Xiangyu~Zhang et~al.,
\newblock ``Deep residual learning for image recognition,'' 2015.

\bibitem{Detectron2018}
Ross Girshick and Ilija~Radosavovic et~al.,
\newblock ``Detectron,'' 2018.

\bibitem{coco-eval}
Tsung-Yi Lin and Michael~Maire et~al.,
\newblock ``Microsoft coco: Common objects in context,'' 2015.

\end{thebibliography}


\end{document}